\pdfoutput=1
\documentclass[11pt]{article}

\usepackage{amsmath}
\usepackage{amssymb}
\usepackage{array}
\usepackage{ACL2023}

\usepackage{enumitem}% http://ctan.org/pkg/enumitem

\usepackage{gb4e}
\noautomath
\usepackage{geometry}
\usepackage{times}
\usepackage{latexsym}
\usepackage{graphicx}
\usepackage{soul}
\usepackage{multicol}
\usepackage{multirow}
\usepackage{subcaption}
\usepackage{listings}
\usepackage{color}
\usepackage{booktabs}
\usepackage[T1]{fontenc}
\usepackage[utf8]{inputenc}

\usepackage{microtype}

\usepackage{inconsolata}

\title{SEC-QA: A Systematic Evaluation Corpus for Financial QA}

\author{Viet Dac Lai, Michael Krumdick, Charles Lovering\\
{\bf Varshini Reddy, Craig Schmidt, Chris Tanner} \\
  Kensho Technologies\\
  \texttt{\{viet.lai,michael.krumdick,charles.lovering,chris.tanner\}@kensho.com} 
}

\definecolor{mygreen}{rgb}{0,0.6,0}
\definecolor{mygray}{rgb}{0.5,0.5,0.5}
\definecolor{mymauve}{rgb}{0.58,0,0.82}

\definecolor{C_BASE}{HTML}{007694}
\definecolor{C_BASE_SOFT}{HTML}{66adbf}
\definecolor{C_BASE_SOFTER}{HTML}{b9e5f0}

\definecolor{C_SECOND}{HTML}{B62699}
\definecolor{C_SECOND_SOFT}{HTML}{e2a8d6}
\definecolor{C_SECOND_SOFTER}{HTML}{f2c7ea}

\definecolor{C_THIRD}{HTML}{00B9E8}
\definecolor{C_FOURTH}{HTML}{0B008F}
\definecolor{C_FOURTH_SOFTER}{HTML}{b9b6e3}
\definecolor{C_BACKGROUND}{HTML}{EDF5FC}

\newcommand{\mtr}[2]{\multirow{#1}{*}{\textbf{#2}}}
\newcommand{\mtc}[2]{\multicolumn{#1}{c}{\textbf{#2}}}
\definecolor{ks1}{HTML}{007694}
\definecolor{ks2}{HTML}{B62699}
\definecolor{ks3}{HTML}{00B9E8}
\definecolor{ks4}{HTML}{0B008F}
\definecolor{commentgreen}{HTML}{459a4b}

\newcommand{\ksa}[0]{\color{ks1}}
\newcommand{\ksb}[0]{\color{ks2}}

\usepackage{xspace}
\newcommand{\dataset}[0]{SEC-QA\xspace}
\newcommand{\fulldataset}[0]{Systematic Evaluation Corpus for Financial QA\xspace}

\begin{document}
\maketitle
\begin{abstract}
The financial domain frequently deals with large numbers of long documents that are essential for daily operations. Significant effort is put towards automating financial data analysis. However, a persistent challenge, not limited to the finance domain, is the scarcity of datasets that accurately reflect real-world tasks for model evaluation. Existing datasets are often constrained by size, context, or relevance to practical applications. Moreover, LLMs are currently trained on trillions of tokens of text, limiting access to novel data or documents that models have not encountered during training for unbiased evaluation. We propose SEC-QA, a continuous dataset generation framework with two key features: 1) the semi-automatic generation of Question-Answer (QA) pairs spanning multiple long context financial documents, which better represent real-world financial scenarios; 2) the ability to continually refresh the dataset using the most recent public document collections, not yet ingested by LLMs. Our experiments show that current retrieval augmented generation methods systematically fail to answer these challenging multi-document questions. 
In response, we introduce a QA system based on program-of-thought that improves the ability to perform complex information retrieval and quantitative reasoning pipelines, thereby increasing QA accuracy.
%RAG performance, thereby increasing the QA accuracy.
% The questions are published at \url{http://github.com/anonymous}

\end{abstract}

\section{Introduction}

Large Language Models (LLMs) have demonstrated impressive capabilities in a wide range of natural language processing (NLP) applications \cite{brown2020language}. 
Even though LLMs were scaled to an unprecedentedly large number of parameters, they still face many issues with 
hallucinations \cite{ji2023survey}, poor reading comprehension \cite{liu2024lost}, and private data leakage \cite{balloccu-etal-2024-leak}. 
To address these issues, Retrieval Augmented Generation (RAG) produces responses by using a few retrieved documents from reliable sources. 
As a result, it offers more dependable answers with fewer hallucinations, even when employing much smaller language models \cite{borgeaud2022improving}. 
Evaluating RAG-based systems is challenging due to the added complexity of the retrieval step \cite{chen2024benchmarking}, and the cascading effect of the upstream retrieval task on the downstream QA task.
Early benchmarks for RAG-based systems mainly focus on simple common-sense questions that can be answered by retrieving a single piece of text 
from a single knowledge source \cite{joshi-etal-2017-triviaqa,dunn2017searchqa}.  Often these knowledge sources were open textual knowledge bases such as Wikipedia \cite{yang-etal-2018-hotpotqa} and WikiHow \cite{deng2020joint}. 

These benchmarks face two data leakage issues. First, they were created from open Internet sources \cite{xue-etal-2021-mt5,penedo2023refinedweb}, e.g., Wikipedia, which is heavily used in LLM pre-training \cite{touvron2023llama}. 
Second, the static benchmarks themselves are leaked to the Internet \cite{balloccu-etal-2024-leak}. 
LLM training datasets subsequently include these benchmarks, leading to inflation and unreliable results on these benchmarks.
% to LLM benchmarking inflation and unreliable results. 
To combat these problems, there have been proposals to keep benchmarks private \cite{mialon2023gaia} or updated regularly \cite{fan2023nphardeval}.

More importantly, in many financial applications, evaluation based on common-sense knowledge and open knowledge sources may not reflect the system's true capabilities \cite{zhang2024raft}. 
The questions in professional use are far more complex, requiring some combination of multi-hop reasoning \cite{yang-etal-2018-hotpotqa}, 
multi-source reference \cite{tang2024multihop}, document-structure reference \cite{saad2023pdftriage}, and collection structure reference. 
Additionally, knowledge bases in a specific domain may be less diverse than open knowledge bases, 
e.g., Wikipedia, leading to poor retrieval performance which in turn has a severe impact on the LLM's response. 
Retrieval in some domain-specific areas may instead require domain knowledge to achieve effective retrieval that has not been captured in open-domain evaluation.

To overcome these issues, this work introduces a framework for designing practical quantitative questions. These questions are much more challenging than existing financial QA tasks such as FinQA\cite{chen-etal-2021-finqa} or TAT-QA \cite{zhu-etal-2021-tat}.
The framework allows us to customize questions at the needed complexity for the target applications with potential variety in question complexity 
including {\it multiple entities/financial periods}, {\it multi-hop reasoning}, {\it document structure}, {\it collection structure},  and {\it multiple outputs}. 
We leverage Internet-accessible document collections, and open tabular databases to create real-world complex quantitative questions in finance. 
% Using this framework, we design a set of practical quantitative questions that are very challenging. 
We evaluate four RAG-based systems and show that RAG systems systematically fail on these carefully designed real-world questions. 
% Moreover, we show that recent LLMs using code generation can leverage financial knowledge to navigate the financial document effectively, leading 
Moreover, we show that recent LLMs can use code to effectively navigate the structure of the document collections; e.g., that earnings per share information is within 10-Ks, and that there are 10-Ks for each company and fiscal year. This leads to drastically improved levels of performance. 
%This experiment shows how the structure of the document collection can be adopted effectively to boost the performance of question-answering. 
Additionally, this framework can be used to dynamically refresh the benchmarks regularly to prevent training data leakage.

The contributions of this paper are:
\begin{itemize}[noitemsep]
  \item A framework (\dataset) for dynamically generating quantitative multi-hop QA tasks for the financial domain from publicly accessible documents and databases.
    \item A set of practical and challenging questions for Quantitative Reasoning in the financial domain that vanilla RAG models systematically fail to answer.
    \item A system that utilizes program-of-thought and the rich structure of the document collection to improve QA performance.
\end{itemize}
% \vspace{-\topsep}

% \begin{itemize}
%     \item A set of practical and challenging questions for Quantitative Reasoning in the financial domain that vanilla RAG models systematically fail to answer.
%     \item A system that utilizes program-of-thought and the rich structure of the document collection to improve QA performance.
%     \item A framework for dynamically generating quantitative multi-hop QA tasks for the financial domain from publicly accessible documents and databases.    
% \end{itemize}

\section{Related Work}

\newcommand{\yes}[0]{$\color{violet}\checkmark$}
\newcommand{\nah}[0]{$\color{red}$}

\begin{table*}[t]
\centering
\resizebox{0.8\linewidth}{!}{
    \begin{tabular}{clcccrrrc}
        \toprule
        & \mtc{1}{\bf Dataset}      
        & \mtc{1}{\bf Multi-Doc} 
        & \mtc{1}{\bf Multi-Hop} 
        & \mtc{1}{\bf Refreshable} 
        & \mtc{1}{\bf $\#$Test} 
        & \mtc{1}{\bf Context} 
        & \mtc{1}{\bf $\#$Docs} 
        & \mtc{1}{\bf Data } \\
        \midrule
        \mtr{4}{\rotatebox{90}{General}} 
        & HybridQA & \yes & \yes   & -  &   3,463   &   2,326   & 44 & Hybrid  \\
        & TriviaQA    & \yes & - & - & 17,210 & 3,760 & 486,956 & Text  \\
        & SearchQA    & - & - & - &  27,248 & 38  & - & Text  \\
        & HotPotQA    & \yes & \yes & - &  7,405 & 928 & 15,519 & Text  \\
        \midrule
        \mtr{5}{\rotatebox{90}{Finance}} 
        & TAT-QA     &  -  & -    & -  &   1,669   &   47   & - & Hybrid  \\
        & MultiHiertt   & -    & -  & - & 1,566  &  1,646  & - & Hybrid \\
        & FinQA      & - & - & - & 1,147    & 628      & - & Hybrid \\
        & ConvFinQA   & - & - & - & 434 & 628 & - & Hybrid\\
        & Multihop-RAG    & \yes & - & - &  2,556 & 1,574   & 609 & Text\\
        \midrule
        & \dataset    & \yes & \yes & \yes & Flexible  & 123,000   & 1,315 & Hybrid\\
        \bottomrule
    \end{tabular}
}
\caption{Comparison of QA benchmarks in the quantitative and finance domain. Refreshable indicates that the dataset can be automatically renewed/generated with a different document set. Hybrid indicates that the context contains both tabular and textual data.}
\label{tab:dataset}
\end{table*}

{\bf Reasoning Capabilities of LLM} has been found in natural language processing and other fields.
Some LLMs exhibit emergent capabilities if they are large enough. 
A simple prompt ``{\it Let's think step by step}'' causes a model to generate solutions with reasoning steps in a chain-of-thought \cite{wei2022chain}. 
More advanced prompting techniques have been discovered which are similar to human reasoning processes 
such as tree-of-thoughts \cite{yao2024tree} and self-verification \cite{weng-etal-2023-large}. 
However, the numerical reasoning of LLMs is still limited, motivating the adoption of programming languages to offload numerical tasks in program-of-thought \cite{chen2022program} 
and program-synthesis \cite{austin2021program}. Reasoning capability is enhanced in additional training on reasoning through human feedback \cite{ouyang2022training}. 
However, LLMs still struggle with many domain-specific tasks such as finance \cite{koncel2023bizbench}. 
% Benchmarking is essential to understand a model's behavior. 
Recent studies have pointed out that many popular benchmarks are contained in LLM pre-training data \cite{riddell2024quantifying}, which causes the inflation of model performance. 
As such, benchmarks are kept private \cite{mialon2023gaia}, or updated regularly \cite{fan2023nphardeval} to mitigate data contamination.

{\bf Document Grounded Quantitative Reasoning} involves numerical extraction and numerical reasoning. 
Previous work in NLP has explored numerical extraction from scientific documents \cite{harper-etal-2021-semeval,elazar-etal-2019-large} 
and financial documents \cite{loukas-etal-2022-finer}. Existing datasets for the financial domain that require quantity extraction include
HybridQA \cite{chen-etal-2020-hybridqa}, TAT-QA \cite{zhu-etal-2021-tat}, 
MultiHierTT \cite{zhao-etal-2022-multihiertt}, FinQA \cite{chen-etal-2021-finqa}, and ConvFinQA \cite{chen-etal-2022-convfinqa}. 
However, these works only involve a small amount of grounding context (e.g., a single page, a single document).

{\bf Multi-Document QA}: TriviaQA \cite{joshi-etal-2017-triviaqa} and SearchQA \cite{dunn2017searchqa} require the model to search over a large collection of documents.
However, the question itself can be answered by reading a few sentences extracted from a single document.  
Some multi-document QA datasets were created for open-ended QA such as summarization (MultiNews \cite{fabbri-etal-2019-multi}, 
WikiHowQA \cite{bolotova-baranova-etal-2023-wikihowqa}) where skimming over given documents and extracting evidential cues from these documents are essential. 
The HotpotQA \cite{yang-etal-2018-hotpotqa} dataset specifically targets multi-hop questions to resolve hidden cross-document reference entities in the questions. 
However, these datasets were collected from open knowledge bases (e.g., Wikipedia), so they have most likely been leaked in LLM pre-training data \cite{touvron2023llama}.

Multihop-RAG \cite{tang2024multihop} propose a multi-hop dataset for financial documents that differs from our work in the following ways: (i) their work studies multi-hop reasoning only with regards to parallel retrieval queries, while we consider both parallel and sequential reasoning steps; (ii) the news documents used to create their dataset do not reflect the real-world sources such as official filings used by financial professionals, that often span hundreds of pages.

%{\bf Financial NLP} non-quantitative tasks have been used to evaluate financial NLP models such as named entity recognition \cite{salinas-alvarado-etal-2015-domain}, 
{\bf Financial NLP} has explored non-quantitative tasks such as named entity recognition \cite{salinas-alvarado-etal-2015-domain}, 
sentiment analysis \cite{malo2014good}, classification \cite{sinha2021impact}, 
question answering \cite{maia2018www}, boundary detection \cite{au2021finsbd}, and entity/event extraction \citet{Lu2023BBTFinCC}. 
% Program synthesis has caught the attention of the financial community thanks to its numerical precise answer and auditability \cite{koncel2023bizbench}.

\section{Framework Construction}
We propose \fulldataset (\dataset), a framework for generating financial Multi Document Questions and Answers (MDQA). We also refer to the questions generated by this framework with the same name, \dataset.

\subsection{Task Definition}
MDQA is defined as follows: A system $S$ is asked a question $q$ with answer $a$. $q$ can be answered by looking in at the document collection $C=\{D_i|1\leq i \leq N\}$. 
Each document consists of several pages $p_{i}=(t_{ij}, c_{ij})$ where $t_{ij}$, $c_{ij}$ are the title and the content, respectively.

% First, we require Second, Requires a dataset collection contains either the metrics in B or the constituent submetrics.

% \begin{figure}[t]
%     \begin{subfigure}{\linewidth}
%         \includegraphics[width=\linewidth]{figures_vr/capiq_pro_table.pdf}
%         \caption{A balance sheet table from Capital IQ Pro}
%         \label{fig:capitaliq_pro_table}
%     \end{subfigure}
%     \begin{subfigure}{\linewidth}
%         \includegraphics[width=\linewidth]{figures_vr/capiq_pro_document.pdf}
%         \caption{A balance sheet table from Capital IQ Pro}
%         \label{fig:capitaliq_pro_document}
%     \end{subfigure}
% \end{figure}

\subsection{Resources}
%Many quantitative datasets in finance are the collection of data from many textual sources such as public reports in both public sectors, e.g., Annual Comprehensive Financial Report (ACFR) published by various governmental entities such as state, city, and department, and private sectors, e.g., annual/quarter financial reports (Form 10-K/10-Q) required by the United State Securities and Exchange Commission (SEC). In this work, we focus on private sector finance where the key financial metrics are provided by market-trusted major Finance Information providers. As such, we assumed their database is highly comprehensive and accurate.
\dataset, a framework for flexibly creating questions for MQDA, requires: 1) Database \textit{T} with values partitioned by variables of interest (e.g., revenue partitioned by company and fiscal year). 2) Document collection \textit{C} that contains the information needed to compute the values within \textit{T}.
% \begin{enumerate}[noitemsep]
%     \item Database \textit{T} with values partitioned by variables of interest (e.g., revenue partitioned by company and fiscal year.)
%     \item Document collection \textit{C} (as defined above)  aligned with \textit{T} such that \textit{C} contains the values within \textit{T} or constituent sub-values. If \textit{T} tracks the documents in \textit{C} that source each value then in addition to question accuracy it is possible to report document and page-level retrieval metrics.
% \end{enumerate}

Specifically, we leverage private-sector financial data from market-trusted sources, ensuring comprehensive and accurate datasets. We collect key financial metrics and their associated documents to create a database (key-value-document table) $T \in (c, y, k, v, d)$, where $v$ represents the value of the key metric $k$ for company $c$ in fiscal year $y$ as mentioned in document $d$. Because \textit{T} tracks the documents in \textit{C} that source each value then in addition to question accuracy it is possible to report document and page-level retrieval metrics.

Our collection comprises 10 metrics for 18 publicly traded companies from the S\&P 500 list from 2010 to 2023. We also collect their annual reports (Form 10-K), quarterly reports (Form 10-Q), and unscheduled event reports (Form 8-K) for the same period. Since the documents are published in HTML format, we convert the documents into PDF(s), and then we parse the PDF(s) into JSON format using a public PDF extraction service
\footnote{\url{https://kensho.com/extract}}.
% \footnote{\url{anonymous.com}}. 
Documents are represented as JSON objects with paragraphs, well-structured tables, and machine-detected titles.

%We also collect their annual reports (Form 10-K), quarterly reports (Form 10-Q), and unscheduled event reports (Form 8-K) for the same period. Since the documents are published in HTML format, we convert the documents into PDF, and then we parse the PDF document into JSON format using a public PDF extraction service.\footnote{anonymous.com} This process results in a clean document (without HTML tags) with rich content including textual paragraphs, well-structured tables, and machine-detected titles.
%In total, our database consists of 1500 documents for a total of 77K pages from 18 companies over a 14-year period.

\subsection{Question Complexity}

Question design is an important step for a successful QA system evaluation. 
In this work, we identify several factors that increase the complexity of a question in the financial domain.

We define an {\bf atomic question} as a question that seeks a single piece of information that can be extracted directly from a single document. Such questions usually involve a single entity for a particular financial period, 
e.g., {\it ``What is the \underline{total revenue} of \underline{Apple Inc.} for the \underline{fiscal year 2022}?''}. 

% However, in financial analysis, questions are usually much more complex. 
% They require extracting multiple evidential information cues, transforming the extracted information, and outputting the answer in various output formats and media. 

However, in financial analysis, questions tend to be much more complex. They require extracting multiple pieces of data, transforming the extracted data, and presenting the answer in various formats. Further, we have observed the following main challenges in multiple-document QA for finance:

% In this work, we provide a framework for easily composing complex quantitative questions from existing high-quality databases. 

\begin{itemize}[noitemsep]
    \item {\bf Parallel Reference} questions require the same kind of information over multiple entities or time periods, e.g.,
     {\it ``What is Intel's revenue growth in the last \underline{5-year period}?''}. This requires extracting information from a few documents. The complexity of parallel reference can be measured by the number of entities/years needed to answer the question.
    \item {\bf Multi-hop Reference} questions require reference resolution of a few implicitly defined entities through some deterministic constraints 
    e.g. {\it ``Show the 5-year stock price history of the \underline{top 5 most valuable companies} in the S\&P 500 index''}. 
    The complexity of multi-hop questions is usually measured by the number of hops needed to answer the question and the complexity of the constraints.
    \item {\bf Structural Reference} involves document structure reference and collection structure reference. Document structure reference refers to a particular section/table/figure in a document \cite{saad2023pdftriage}. Collection structure reference refers to a subset of documents in the collection (e.g., {\it ``recently quarterly filings'', ``their earning calls''}), narrowing the document search space.
    \item {\bf Multi-Output} questions expect multiple values 
    (e.g., {\it ``Analyze the financial performance of the top 3 competitors of Amazon.com, Inc. in terms of revenue for the last 4 quarters, by computing their \underline{revenue growth}, \underline{gross margin}, and \underline{operating margin}.''}. To our knowledge, this type of question has not been addressed in previous work.
\end{itemize}

\subsection{Question Design}

{\bf Question Template}: % Question design is an important part of the benchmark. 
Previous work in QA has tackled some challenges in QA such as multi-hop \cite{yang-etal-2018-hotpotqa}, and document structure \cite{saad2023pdftriage} to target some specific complex question types.
Our framework allows many question types including parallel reference, multi-hop reference, document structure, collection structure reference, and multi-output questions. 
Moreover, due to the rich information in the database, we can design questions to require finance domain jargon (e.g., using a company's stock symbol to refer to the company) and language regularities (e.g., omitting financial periods if asking for the latest figures). 
Using this framework, we can flexibly opt for different combinations of complexity (e.g., parallels and multi-hop in the same question).

{\bf Template filling}: The question template can be filled semi-automatically through a simple rule-based system that randomly picks the entities, 
metrics, and periods from the database. 
This allows control over the complexity, quality, and distribution of the generated questions and answers.

\begin{table*}[t]
\resizebox{\linewidth}{!}{
    \begin{tabular}{l|l}
        \toprule
        \multicolumn{1}{c|}{\bf Function} & \multicolumn{1}{c}{\bf Description} \\
        \midrule
        {\ksa select\_document}({\ksb stock\_symbols}, {\ksb form\_types}, {\ksb fiscal\_years}) & Select a few documents given some filters \\
        \midrule
        {\ksa retrieve\_relevant\_pages}({\ksb text\_query}, {\ksb documents}) & Retrieve top-$k$ pages from the given documents \\
        \midrule
        {\ksa extract\_value}({\ksb text\_query}, {\ksb pages}) & Extract a value and its corresponding multiplier from the given pages\\
        \bottomrule
    \end{tabular}
}
    \caption{List of helper functions that the CodeGen systems can use}
    \label{tab:functions}
\end{table*}

\section{Experiment}

This section presents our findings across three use cases of the \dataset framework.
% This section analyzes the multi-document QA pipeline, presenting our findings across three use cases.
 %, and the findings from these experiments. 

\subsection{QA Systems}

We evaluate 4 systems with different characteristics 
for a broad understanding of MDQA in finance:

\begin{itemize}[noitemsep]
    \item {\bf Vanilla RAG}: To demonstrate the challenge of answering complex questions, 
    we employed a simple retrieval-based system with direct generation.
    \item {\bf Multi Query RAG}: leverages multiple queries for a given input question.
    \item {\bf CodeGen+PageR}: % Program synthesis has become a common approach for numerical reasoning in finance due to the strict requirements of calculation precision that currently LLMs cannot  reliably achieve.
    We employ an LLM to generate code that makes use of two helper functions: {\it ``retrieve\_relevant\_pages''} retrieves $k$ pages from the whole collection;
    {\it ``extract\_value''} calls a prompted LLM to extract the value from the given retrieved pages. This system allows an LLM to decompose a complex question into atomic questions as well as make a full plan of how to answer this question. In turn, it allows us to examine the planning capability of the LLM.
    \item {\bf CodeGen+DocS+PageR}: Financial document collections track meta information for each document like company, fiscal year, and form type. This can be used to reduce the retrieval search space. 
    Beyond the two functions {\it ``retrieve\_relevant\_pages''} and {\it ``extract\_value''}, 
    we introduce a {\it ``select\_document''} function
    that filters the document collection based on meta information such as the company stock symbol and fiscal year. Critically, a document can be selected and then queried for pages.
\end{itemize}

% \begin{itemize}[noitemsep]
%     \item {\bf Vanilla RAG}: This system employs a straightforward retrieval-based approach with direct generation. It generates an answer for a question using the document retrieved from the entire corpus.
%     \item {\bf Multi Query RAG}: This system allows multiple queries for a given input question to generate the answer, thereby enhancing retrieval accuracy.
%     \item {\bf CodeGen+PageR}: In this setting, we employ an LLM to generate code utilizing two helper functions: {\it ``retrieve\_relevant\_pages''}, which retrieves \( k \) pages from the entire collection, and {\it ``extract\_value''}, which calls a prompted LLM to extract the value from the given retrieved pages. This system enables the LLM to decompose a complex question into atomic questions, thereby evaluating the planning capacity of the prompted LLM.
%     \item {\bf CodeGen+DocS+PageR}: Financial document collections track meta information for each document, such as company, fiscal year, and form type, which can be used to reduce the retrieval search space. In addition to the previous two functions, we introduce a document selection function, {\it ``select\_document''}, that filters the document collection based on the company ticker and fiscal year. Thus, allows a document to first be selected and then queried for specific pages.
% \end{itemize}

We use OpenAI's Ada as the neural embedding for retrieval and GPT4 (gpt-4-1106-preview) as the LLM for all our experiments. We use the same three questions as the exemplars for the CodeGen system, varying only the available helper functions. We test different numbers of retrieved pages ($k\in[4,128]$) and report the best performance.

\subsection{Evaluation}

Due to the complexity of the pipeline and models we used, we evaluate the model's performance in 3 stages: document retrieval, page retrieval, and question answering. 
For document retrieval and page retrieval, we report the Precision@K, Recall@K, and F1@K. 
For systems that only involve page retrieval, we report their document retrieval performance based on the document signature of the retrieved pages. 
Since many numbers in financial reports are rounded to various levels (thousands, millions, billions), using exact matches for automatic answer scoring is challenging. Therefore, we accept an answer if its value is within a 1\% margin of error from the golden answer.
% For question-answering tasks, because many numbers in financial reports are rounded at some level (thousands, millions, billions) 
% depending on the context and for the sake of readability, it is hard to use the exact match in automatic scoring the answer.
% Instead, we accept an answer if its value is within 1\% margin of error of the golden answer.

\subsection{Single Value Extraction Task}
\label{sec:value_extraction_task}
We begin with a simple extraction MDQA task that requires a model to retrieve an exact numeric span (e.g., 1234.5) and unit value (e.g., millions) from a document. 
For this use case, we use \dataset to generate questions based on the following templates: \eqref{ve1}
is usually used for the latest update of a metric; \eqref{ve2} is used for the previous financial periods (e.g., years and quarters). 
% \begin{quote}
%     \it
%     VE-1: What is \{company\}'s \{metric\}?\\
%     VE-2: What is \{company\}'s \{metric\} in \{year\}?
% \end{quote}

\begin{exe}
   \ex \label{ve1} \textit{What is \{\textbf{company}\}'s \{\textbf{metric}\}?} 
   \ex \label{ve2} \textit{What is \{\textbf{company}\}'s \{\textbf{metric}\} in \{\textbf{year}\}?}
\end{exe}

Financial analysts use this language regularly, in which the year is omitted from the question. 
As such, a model must catch that regularity to identify the correct value term to extract.
To confirm the existence of the metric in the document, we remove the questions whose answers can not be found in the document with simple string matching.

\begin{figure}[t]
    \includegraphics[trim={0 0.5cm 0 0.8cm}, width=\linewidth]{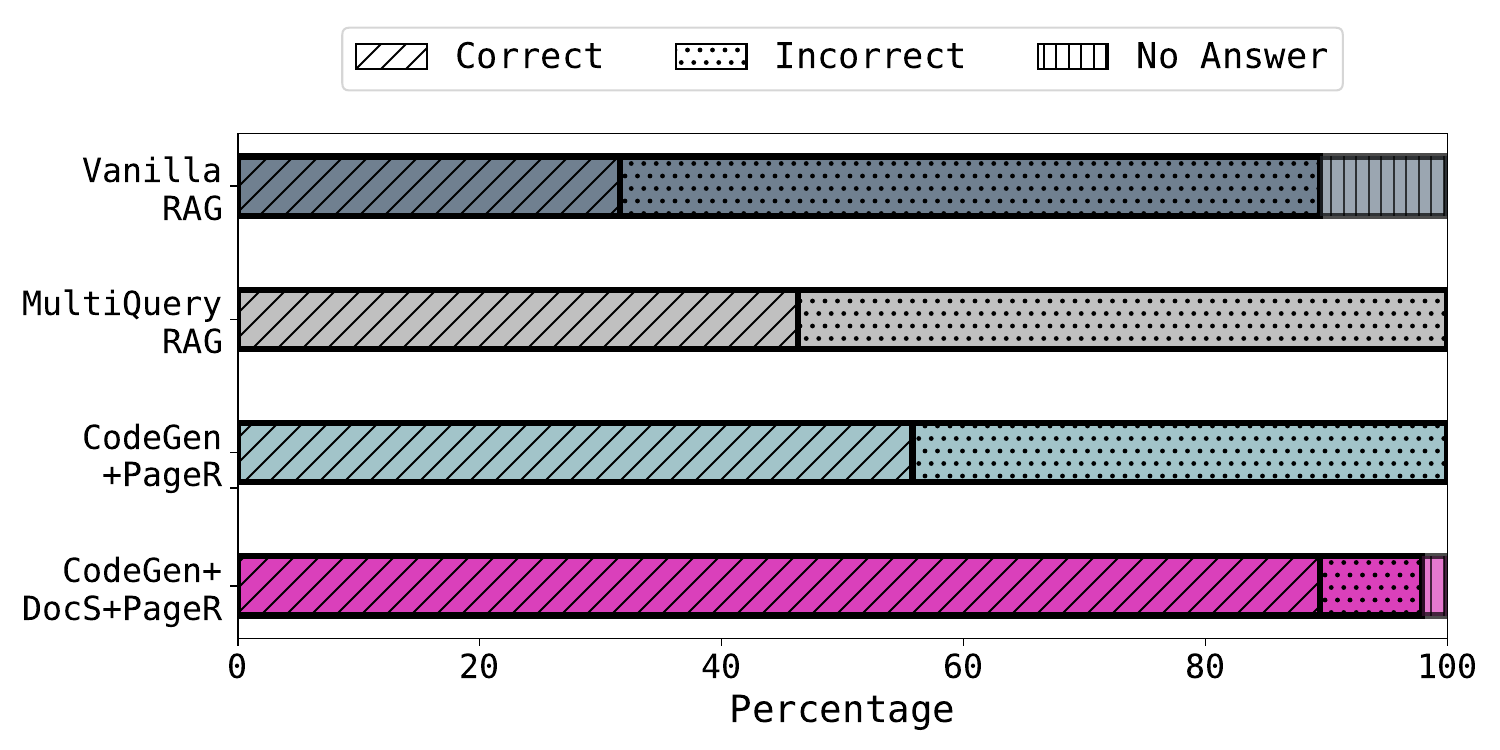}
    \caption{Performance of the simple value extraction}
    \label{fig:result_value_extraction}
\end{figure}

Figure \ref{fig:result_value_extraction} shows the performances of the models of this single-document value extraction task. 
CodeGen+DocS+PageR system performs best, with 89.5\% accuracy. 
On the other hand, the CodeGen+PageR and Vanilla-RAG systems lag behind with 31.6\% and 55.8\%  accuracy. 
% Comparing the Vanilla-RAG and CodeGen+PageR, due to the simplicity of this 
% Value Extraction task, there are not many differences in how the task is done between these two models. 
% However, CodeGen+PageR still yields higher accuracy than Vanilla-RAG. 
Comparing the two CodeGen models, we can see that the CodeGen w/ DocS outperforms the variant w/o DocS. 
This suggests that document selection is important for financial questions. This result also shows that a general neural-based document retrieval struggles to cope with the demanding requirements for retrieval in finance. Part of the reason the 
 neural-based retrieval struggles is because of the structure of the public filings in the financial domain. 
Many documents are very similar to each other, especially documents for the same company. Long identical phrases are often used for multiple years. Without document selection, LLMs end up having to process irrelevant pages collected from previous years. Section \ref{sec:case_study} shows an example of models without document selection.
This experiment shows there is a large performance difference for different pipeline settings.

\subsection{Compound Value Extraction Task}
Compound or high-order metrics appear frequently in financial analysis. 
They are usually computed based on a few other metrics reported in the public filings. 
As such, being able to answer questions with compound metrics is crucial to automating the pipeline of financial analysis. 
Previous work considers these questions as {\it NULL} \cite{tang2024multihop}, skipping the question.
In real-world applications, a model should provide its best estimation based on the provided information, ideally with an explanation to justify the estimation.

To do this, we design a set of questions with compound metrics, such as Revenue Per Employee (RPE) using the same Templates \ref{ve1}, \ref{ve2}.
Some companies report these metrics in their filings, so we only consider metrics that cannot be easily found with string matching. 
Thus, the model must be able to understand the formula to calculate to answer correctly. 
In total, we generated 24 of these questions.
Because the value is not explicitly in the text, these questions require both \textit{a priori} knowledge of the metrics and additional reasoning steps to compute the metric. Therefore, we expect these questions to be more difficult.
% We hypothesize that since the value is not explicitly expressed in the text, it is hard for the model to calculate or estimate the value precisely. 
% We also create a variant of the questions that the formula description is included in the question (See Appendix \ref{}). 

Table \ref{tab:result_compound_value_extraction} shows the performance of the model on this question set. Performance for all tested systems on these Compound Value Extraction questions is significantly lower than the Single Value Extraction questions (as presented in Section \ref{sec:value_extraction_task}).

% \begin{equation*}
%     \footnotesize
%     \begin{aligned}
%         \text{RPE} =& \frac{\text{Total Revenues}}{\text{Total Full-time Employees}}\\
%         \text{Total Debt} =& \text{Short-term Borrowings}\\
%         &+ \text{Current Portion of Long-Term Debt}\\
%         &+ \text{Current Portion of Leases}\\
%         &+ \text{Long-Term Debt}\\
%         &+ \text{Long Term Leases} ...\\
%         % &+ \text{Finance Div. Debt Current}\\
%         % &+ \text{Finance Div. Debt Non-Curr.}\\
%     \end{aligned}
% \end{equation*}

% The performance on compound metric questions is much lower than on non-compound metric questions . 

% All the models struggle with this task and the performance drops significantly.

\begin{table}[]
    \centering
    \small
    \begin{tabular}{lc}
        \toprule
        \bf System & \bf Correct (\%) \\
        \midrule
        Vanilla RAG & 8.3 \\
        Multi Query RAG & 16.7 \\
        CodeGen+PageR & 37.5\\
        CodeGen+DocS+PageR & 33.3\\
        \bottomrule
    \end{tabular}
    \caption{Performance on compound value extraction.}
    \label{tab:result_compound_value_extraction}
\end{table}

% More importantly, comparing the performance without and with the formula description, we also see a drop in the accuracy, 
% indicating that the description does not provide valuable information for the model to calculate the compound metrics.

For this task, we are able to investigate the performance at the sub-metric level thanks to the sub-metric data in the database. The Vanilla RAG and Multi Query RAG models do not extract value at the sub-metric level, so we omit these models from this analysis. 
We find that CodeGen-based models systematically query each sub-metric (e.g., Long-Term Debt, and Long-Term Leases). 
However, once the model unrolls the main metric (e.g., total debt) into sub-metrics, many sub-metrics are missing from the document due to two reasons: (1) the sub-metrics are also compound metrics and (2) some sub-metrics do not apply to some companies. 
These lead to a false extraction or duplication (Long-Term Debt being extracted twice for Long-Term Debt and Long-Term Leases).

% \begin{figure}
%     \centering
%     \includegraphics[width=\linewidth]{figures_vr/result-compound-metrics.pdf}
%     \caption{Caption}
%     \label{fig:result_compound_metrics}
% \end{figure}

This test highlights the difficulty of the value extraction task in the financial domain. 
This also shows how we can easily use our framework to customize the test without data annotation.

\subsection{Multi-Document QA Task}\label{subsec:mdqa}

To measure the performance on more complex question, we design a set of question templates that refer to metrics of different years and companies as follows:
% lets you refer to ex
% \setcounter{exx}{0}
\begin{exe}
   \ex \textit{How much common dividends did {\bf\{company\}} pay in the last {\bf \{num\_year\}} years in US dollars? }
   \ex \textit{ What is the percentage difference of {\bf \{company1\}}'s {\bf \{metric\}} compared to that of {\bf \{company2\}}?}
   \ex \textit{ What is {\bf \{company\}}'s overall revenue growth over the last {\bf \{num\_year\}}-year period?} 
   \ex\label{md4}\textit{Among {\bf \{company\_names\}}, what is the {\bf \{metric2\}} of the company that has the highest {\bf \{metric1\}}?}
\end{exe}
% \begin{exe}
% \it
%    \exi{MD-1} How much common dividends did {\bf\{company\}} pay in the last {\bf \{num\_year\}} years in US dollars? 
%    \exi{MD-2} What is the percentage difference of {\bf \{company1\}}'s {\bf \{metric\}} compared to that of {\bf \{company2\}}?
%    \exi{MD-3}  What is {\bf \{company\}}'s overall revenue growth over the last {\bf \{num\_year\}}-year period? 
%    \exi{MD-4}\label{md4}  Among {\bf \{company\_names\}}, what is the {\bf \{metric2\}} of the company that has the highest {\bf \{metric1\}}?
% \end{exe}
% \eqref{md4}
% \begin{quote}
%     \it 
%     MD-1: How much common dividends did {\bf\{company\}} pay in the last {\bf \{num\_year\}} years in US dollars? \\
%     MD-2: What is the percentage difference of {\bf \{company1\}}'s {\bf \{metric\}} compared to that of {\bf \{company2\}}? \\
%     MD-3: What is {\bf \{company\}}'s overall revenue growth over the last {\bf \{num\_year\}}-year period? \\
%     MD-4: Among {\bf \{company\_names\}}, what is the {\bf \{metric2\}} of the company that has the highest {\bf \{metric1\}}?
% \end{quote}

\begin{figure}
    \includegraphics[width=\linewidth]{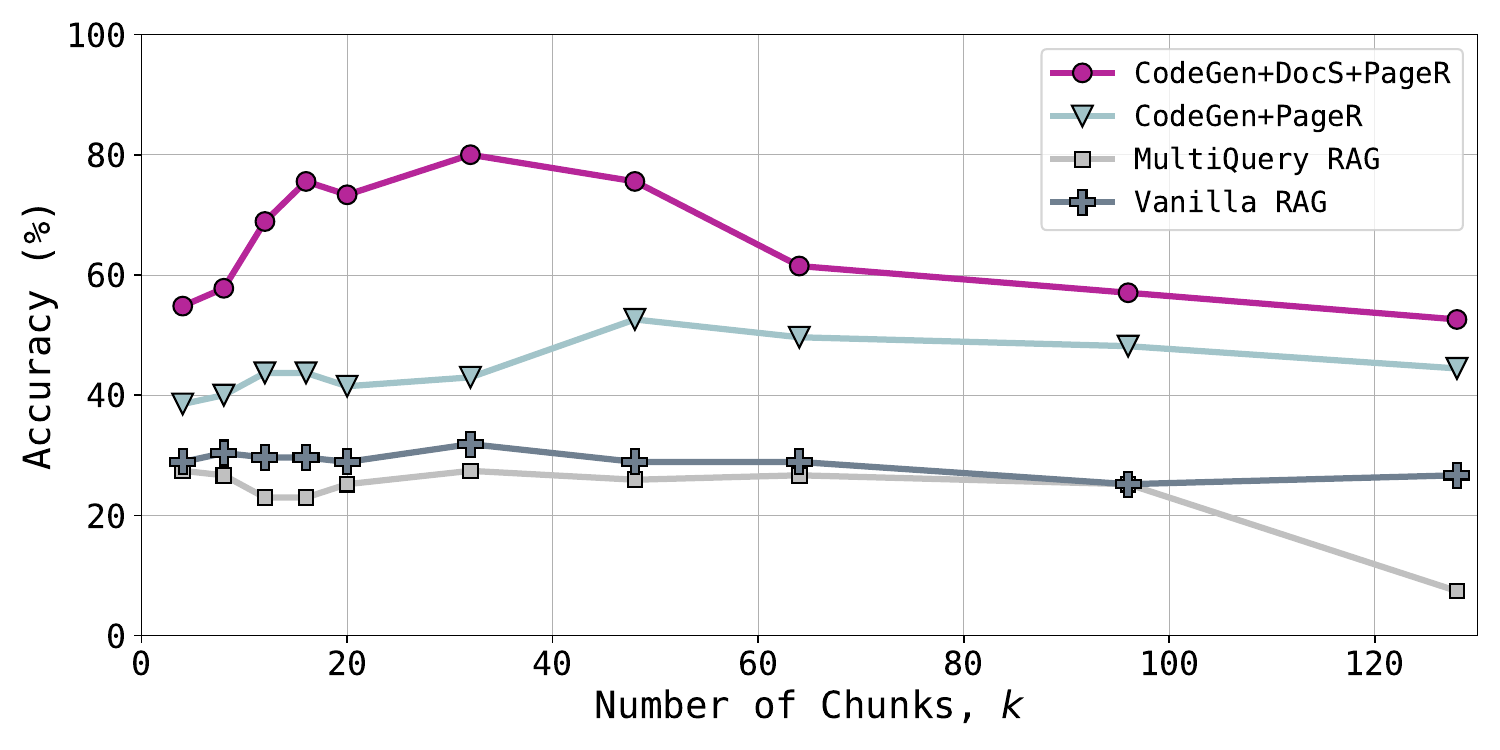}
    \caption{Performance of systems on Multi-Doc QA.}
    \label{fig:result_multi_doc_chunk_performance}
\end{figure}

Figure \ref{fig:result_multi_doc_chunk_performance} shows the accuracy
for the three models on this task with different numbers of retrieved pages $k$. 
We can see that CodeGen+DocS+PageR outperforms the other models with a high margin, 
correctly answering 108 of 135 questions ($80\%, k=32$) compared to 71 with CodeGen+PageR ($52\%, k=48$) and only 41 with Vanilla RAG ($30\%,k=32$). 
Notably, CodeGen+DocS+PageR outperforms the other two models with a very low number of retrieved pages ($k=4$) with 74 correct answers ($55\%$). 

More importantly, CodeGen models are more responsive to the number of retrieved pages. 
CodeGen+DocS+PageR  performance improves rapidly when the number of retrieved pages $k$ increases from 4 to 32, 
whereas CodeGen+PageR improves at a lower rate and Vanilla RAG barely improves. 
This suggests that the Vanilla RAG pipeline is bottle-necked at the retrieval step, which we analyze in depth in Section \ref{sec:factors}.

%%%%%%%%%%%%%%
% SECTION 5
%%%%%%%%%%%%%%

\section{Discussion}

\subsection{System Bottlenecks}
\label{sec:factors}

Previous sections highlight how additional retrieval capabilities improve LLM performance on MDQA. In this section, we perform further analysis to identify the main performance bottleneck across the different systems. We reuse the Template \ref{md4}  which has both parallel and multi-hop references. Beyond document/page retrieval and task performance, we also compute the coefficient of determination $R^2$ between these performances. % template (Section \ref{subsec:mdqa})

\begin{figure}[t]
        \centering
    \begin{subfigure}{0.8\linewidth}
    \includegraphics[width=\linewidth]{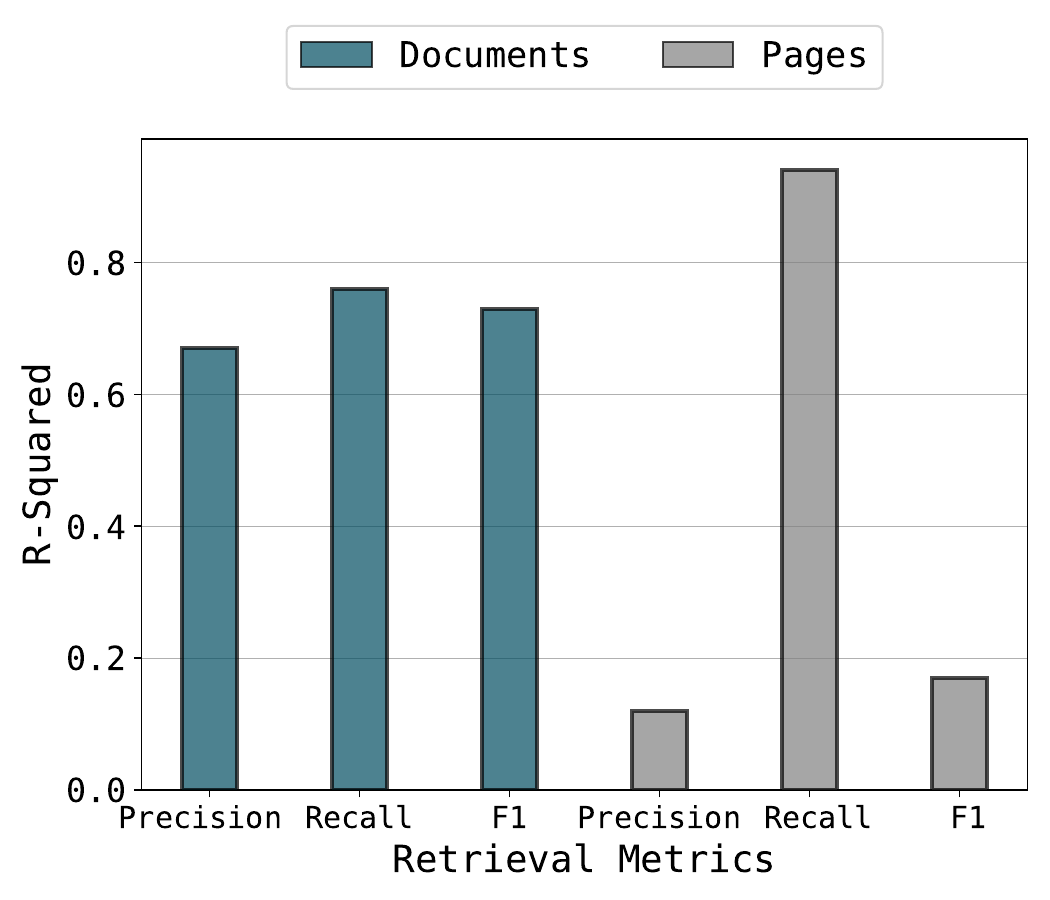}
        \caption{$R^2$ between accuracy and retrieval metrics.}
        \label{fig:result-r-square-metric}
    \end{subfigure}
    \begin{subfigure}{\linewidth}
        \centering
        \includegraphics[width=\linewidth]{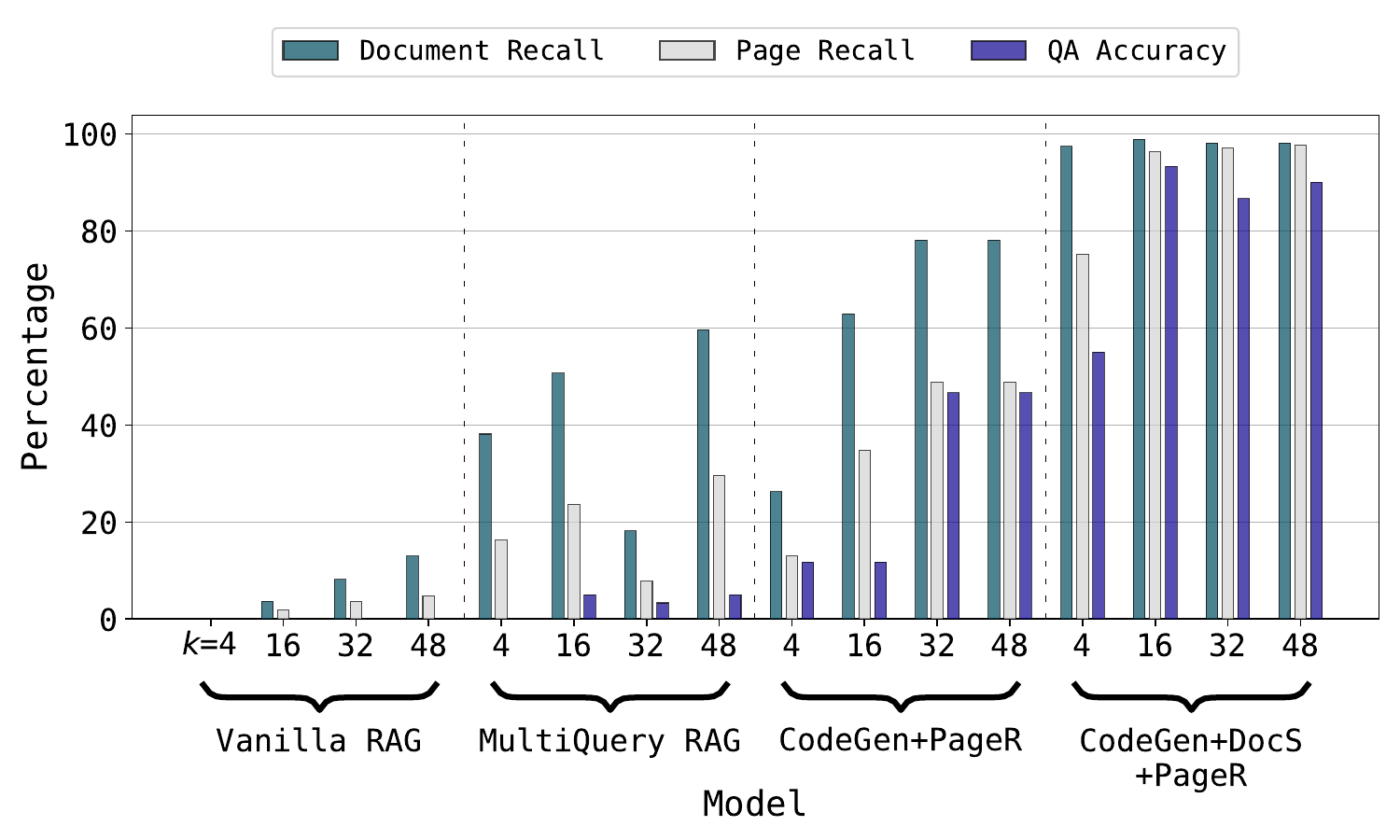}
        \caption{Document-level recall, page-level recall, and QA accuracy of all examined systems with varying $k$ values.}
        \label{fig:result-extraction}
    \end{subfigure}
    \caption{Retrieval Performance}
\end{figure}

Figure \ref{fig:result-r-square-metric} shows the $R^2$ values between the system accuracy and retrieval metrics at both document and page levels. 
Recall has the highest $R^2$ values, with 0.94 and 0.76 for page and document levels respectively, suggesting that recall performance is crucial to the accuracy of the multi-document QA performance. 
On the other hand, page-level precision has a low $R^2$ value of 0.12, indicating a weak correlation with overall performance.

From Figure \ref{fig:result-extraction}, we observe that the QA accuracy of a system is directly proportional to document and page-level recall scores, supporting our previous claim about the correlation between recall and accuracy. Specifically, we note that a model's accuracy aligns more closely with the page-level recall score. The Vanilla RAG model lags behind the CodeGen-based models in terms of accuracy. However, when it is allowed to use multiple retrieval queries (Multi Query RAG), we observe an improvement in retrieval performance, which consequently leads to an increase in QA accuracy.

% From Figure \ref{fig:result-extraction}, we find that the QA accuracy of the systems is directly proportional to document and page level recall scores, supporting our previous claim about the correlation between recall and accuracy. Specifically, we observe that a model's accuracy aligns better with the page-level recall score. 
% The vanilla RAG model lags behind the other CodeGen-based models in model accuracy. 
% % We attribute this to the complexity of the task where the task requires retrieval of textual context from multiple documents.  Using a single retrieval query might yield a sub-optimal retrieval result. 
% In contrast, when this model is free to use multiple retrieval queries (Multi Query RAG), 
% we observe an increase in retrieval performance, thereby leading to an improvement in QA accuracy.

% The best model is the CodeGen+DocS+PageR, where the rule-based document retrieval shows its effectiveness in retrieving documents, 
% yielding a very high document-level recall compared to the CodeGen without DocS.
CodeGen+PageR has a higher recall and QA performance compared to Multi Query and Vanilla RAG. This is due ability of the CodeGen models to break down complex questions into atomic ones and systematically retrieve pages based on the atomic questions. CodeGen+DocS+PageR is observed to be the best model. We attribute this to the addition of the rule-based document selection step, which effectively retrieves relevant documents, thus improving document-level recall.

Overall this experiment shows that our dataset and framework provide us a useful tool to examine multi-document QA in detail. This gives us a better signal for future improvement of the pipeline.

\subsection{Case study}
\label{sec:case_study}

\begin{table*}[t]
\centering
\resizebox{0.8\linewidth}{!}{
   \begin{tabular}{p{1.5cm}|c|l}
      \toprule
      \bf Model & \bf Matched &  {\bf Question:} \it ``What is Adobe's Total Employees reported in 2022?'' \\
      \midrule
         & N & Page: 15 Form type: 10-K; Company: adobe; Fiscal year: \st{\bf 2015}; Period end date: 2015-11-27 \\
        \bf Vanilla & N & Page: 15 Form type: 10-K; Company: adobe; Fiscal year: \st{\bf 2014}; Period end date: 2014-11-28 \\
        \bf RAG     & N & Page: 29 Form type: \st{\bf 8-K}; Company: adobe; Fiscal year: \st{\bf 2020}; Period end date: 2020-12-07 \\
         & N & Page: 35 Form type: 10-K; Company: adobe; Fiscal year: \st{\bf 2012}; Period end date: 2012-11-30 \\
        \midrule
    
        \bf Multi & N & Page: 15 Form type: 10-K; Company: adobe; Fiscal year: \st{\bf 2015}; Period end date: 2015-11-27 \\
        \bf Query & N & Page: 15 Form type: 10-K; Company: adobe; Fiscal year: \st{\bf 2014}; Period end date: 2014-11-28 \\
        \bf RAG & N & Page: 35 Form type: 10-K; Company: adobe; Fiscal year: \st{\bf 2012}; Period end date: 2012-11-30 \\
        & N & Page: 18 Form type: 10-K; Company: adobe; Fiscal year: \st{\bf 2020}; Period end date: 2020-11-27 \\
      \midrule
         & N & Page: 15; Form type: 10-K; Company: ADBE; Fiscal year: \st{\bf 2015}; Period end date: 2015-11-27 \\
         \bf CodeGen  & N & Page: 18; Form type: 10-K; Company: ADBE; Fiscal year: \st{\bf 2023}; Period end date: 2023-12-01 \\
         \bf + PageR  & N & Page: 16; Form type: 10-K; Company: ADBE; Fiscal year: \st{\bf 2021}; Period end date: 2021-12-03 \\
         & N & Page: 15; Form type: 10-K; Company: ADBE; Fiscal year: \st{\bf 2014}; Period end date: 2014-11-28 \\
      \midrule
       \bf CodeGen  & N & Page: 16; Form type: 10-K; Company: ADBE; Fiscal year: 2022; Period end date: 2022-12-02 \\
       \bf + DocS  & N & Page: 2; Form type: 10-K; Company: ADBE; Fiscal year: 2022; Period end date: 2022-12-02 \\
       \bf + PageR  & Yes & Page: 15; Form type: 10-K; Company: ADBE; Fiscal year: 2022; Period end date: 2022-12-02 \\
         & N & Page: 38; Form type: 10-K; Company: ADBE; Fiscal year: 2022; Period end date: 2022-12-02 \\
      \bottomrule
   \end{tabular}
}
\caption{Case study on the inclusion of the page containing the golden answer within the top 4 retrieved pages for the question, \textit{"What is Adobe’s total number of employees reported in 2022?"}}
\label{tab:case_study}
\end{table*}

% Table \ref{tab:case_study} shows the top 4 retrieved pages of the examined models for the question \textit{What is Adobe’s Total Employees reported in 2022?}. The CodeGen+DocS+PageR is the only system that retrieves the page with the golden answer, while other RAG systems and CodeGen+PageR fail. The Vanilla RAG manages to retrieve 10-K pages belonging to Adobe three out of four times, but gets the fiscal year wrong. Multi Query RAG and  CodeGen+PageR only get the fiscal year wrong. This case study shows how financial documents can confuse the modern retrieval systems easily.

Table \ref{tab:case_study} shows the top 4 retrieved pages across QA systems for \textit{"What is Adobe’s total number of employees reported in 2022?"}. Only CodeGen+DocS+PageR  successfully retrieves the page containing the golden answer. Although Vanilla RAG retrieves Adobe's 10-K pages 3 of 4 times, the fiscal year is consistently wrong. Multi Query RAG and CodeGen+PageR also retrieve the wrong fiscal year. This shows how financial documents can easily confuse modern retrieval systems.

\subsection{Stability Tests}
\label{sec:stability_test}

One of the main objectives of this work is to provide a robust benchmark that prevents performance inflation through data leaks by dynamically generating QA pairs using the latest financial documents. This raises the question of whether evaluation scores on a new benchmark version are comparable to those on former versions. To measure the benchmarking consistency, we construct five distinct versions of the dataset using data from different years (2019 to 2023) while keeping the question templates and the set of companies constant to maintain a consistent difficulty level.

The accuracy of the models on different year-based sets varied marginally with small standard deviation ($1.3\%<\sigma<2.0$\%). This shows that we can reliably compare the performance of models between different versions of the benchmark. 
We show the detailed result in Figure \ref{fig:stability_test_year} (Appendix \ref{app:stability_test}).

\subsection{Execution Cost}

% \begin{table}
% \centering
% \small
% \begin{tabular}{lcc}
%     \toprule
%     \bf Model & \bf Retrieval & \bf LLM \\
%     \midrule
%     \bf Vanilla RAG & 1.00 & 1.00 \\
%     \bf CodeGen+PageR & 3.32 & 3.29 \\
%     \bf CodeGen+DocS+PageR & 3.89 & 3.89 \\
%     \bottomrule
% \end{tabular}
% \caption{Average number of success retrieval calls and LLM calls used by the systems. }
% \label{tab:query_count}
% \end{table}

\begin{figure}[t]
    \centering
    \includegraphics[width=0.7\linewidth]{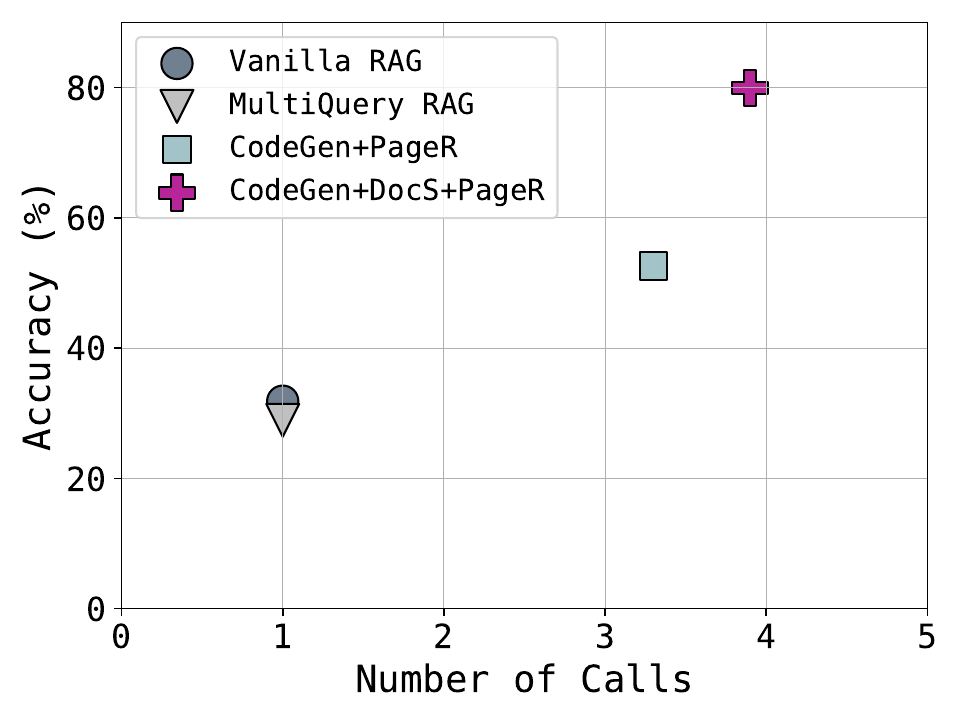}
    \caption{Average number of LLM calls used by the systems in comparison with accuracy.}
    \label{fig:call_vs_accuracy}
\end{figure}

 While code generation-based systems offer superior performance, these models have a higher operational cost and latency. We compute the average number of LLM calls across the four examined systems. From Figure \ref{fig:call_vs_accuracy}, CodeGen systems require approximately four times more calls than Vanilla and Multi Query RAG systems, which can potentially quadruple operational costs and latency. One proposed solution to mitigate latency is to parallelize LLM calls. However, the iterative nature of multi-hop questions poses a challenge to effective parallelization strategies.

\subsection{Automatic Question Design}

Thanks to the advancement in code generation, in principle an LLM with access to a database can generate both questions and code necessary to answer the questions. 
To do that, one can prepare available resources such as documents and functions to access the database and static variable names for relevant entities (e.g. company name, metric name); next, prompt the LLM to generate questions and code that solve the question using the provided functions. 
Once the LLM generates questions and answers, the generated code is executed to obtain the answer. 
After that, one can hire a financial expert to verify the question, the code, and the answer. Notably, the challenge remains at test time as that model will \textit{not} have access to the database $T$, only the document collection $C$.

The benefit of using this method is the potentially higher diversity in question sets. It also provides the code that is used to solve the question. However, this method still requires manual inspection of the question, the code, and the alignment between the question and the code, which are not trivial. In this work, we did not use this method to generate any of the questions used above.

\section{Conclusion}

We introduce \dataset which we leverage to create questions that current RAG approaches consistently fail to answer. This framework can be used to dynamically generate complex practical questions grounded in the financial domain.
Our study highlights the challenges posed by retrieval models in handling multi-doc long-context questions and explores strategies to address these bottlenecks. Furthermore, we propose a method based on program-of-thought and RAG designed to enhance retrieval and downstream performance compared to conventional RAG systems.

\clearpage

\section*{Limitation}

This paper assumes the existence of a collectible set of documents, a tabular dataset of financial metrics, and a method to map these financial metrics to the documents.
We currently explore databases in the private sector, where public reports are heavily regulated, making it relatively straightforward to align the documents with the dataset.

However, in the public sector, reports often vary significantly due to inconsistencies in reporting standards.
As a result, finding a collection of documents, a corresponding dataset and their alignment is more challenging. For instance, our attempts with the US state government's Annual Comprehensive Financial Report (ACFR) and the US Annual Survey of State Government Tax Collections published by the US Census have proven extremely difficult to reverse-engineer into a usable dataset.

\section*{Ethical Consideration}

This dataset was generated automatically from an existing financial database without any involvement of human annotators. Although the CodeGen systems demonstrate significant performance improvements, we do not recommend using them as a replacement for traditional financial analysis tools and financial advice.

\bibliography{anthology.2016.2018,anthology.2018.2020,anthology.2020.2022,anthology.2022.2024,custom}
\bibliographystyle{acl_natbib}

\clearpage

\appendix

\section{Stability Test}

\begin{figure}[!h]
        \centering
        \includegraphics[width=\linewidth]{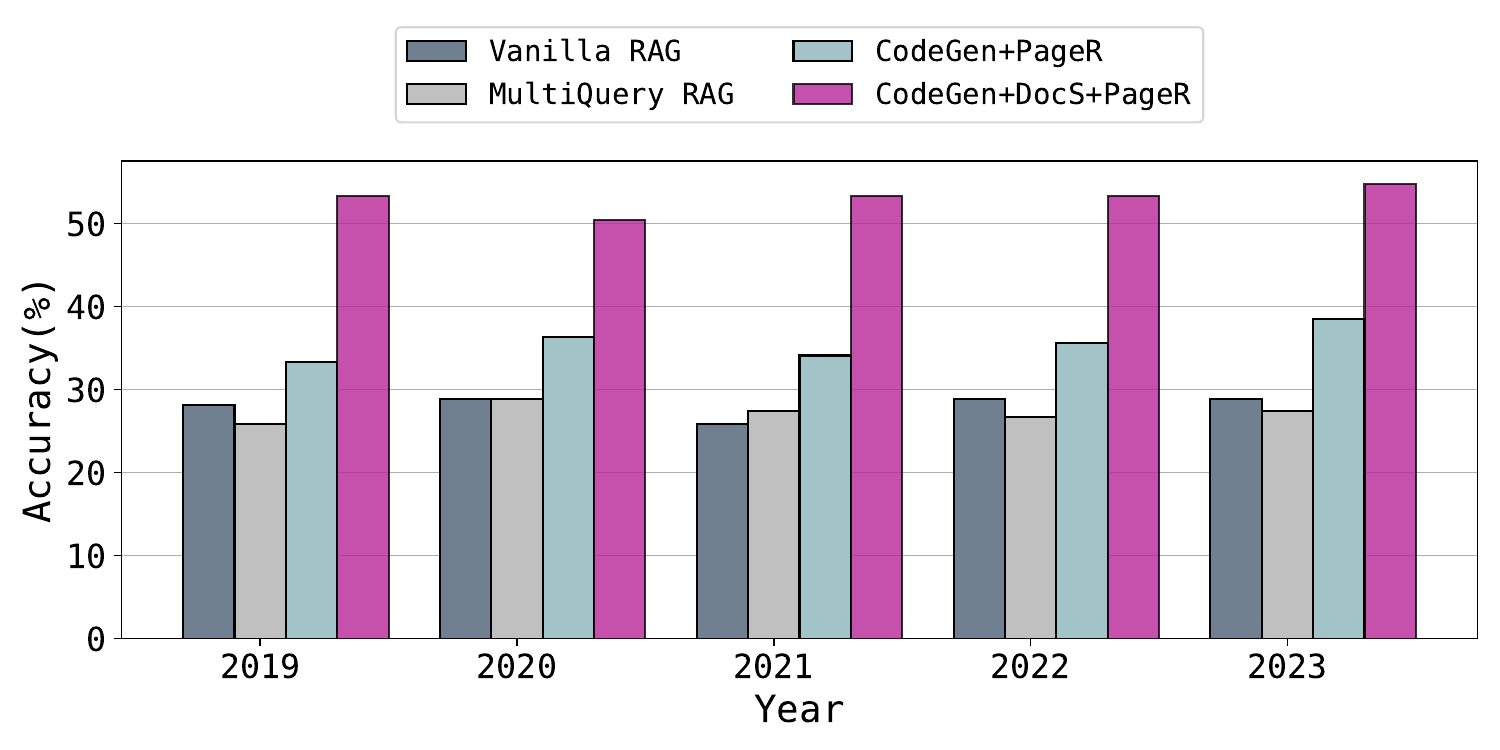}
        \caption{Performance of models on different variants of the same dataset.}
        \label{fig:stability_test_year}
\end{figure}

Figure \ref*{fig:stability_test_year} shows the detailed stability test over the year of the same question complexity mentioned in Section \ref{sec:stability_test}.

\label{app:stability_test}

\section{Model Details}

Throughout the whole study, we use OpenAI's Ada Embedding to encode texts for retrieval.
We use the \texttt{Langchain} implementation of vector-based retrievers (i.e., vanilla and multi-query retrievers).
We use GPT-4 (gpt-4-0125-preview) as the LLM for both the question answering in RAG-based systems and the code generation and value extraction in CodeGen-based systems.

\section{Example of code generated by the CodeGen}
\label{sec:appendix}

Figure \ref{fig:codegen_code_example} shows an example of code generated by the CodeGen+DocS+PageR system.
Figure \ref{fig:codegen_prompt} and \ref{fig:codegen_few_shot_demonstrations} shows the prompt and few-shot demonstrations used to generate code for the CodeGen+DocS+PageR systems.

\begin{figure*}
\begin{lstlisting}[language=Python]
# Among Honeywell, Caterpillar, Pfizer, PepsiCo, and Boeing, 
# What is the Total Revenues of the company that has the lowest Total Debt?

from functions import (
  select_documents, 
  retrieve_relevant_pages, 
  extract_value
)

# Step 1: Identify the total debt for each company
companies = ["Honeywell", "Caterpillar", "Pfizer", "PepsiCo", "Boeing"]
# Assuming these are the correct stock symbols
stock_symbols = ["HON", "CAT", "PFE", "PEP", "BA"]  
total_debts = {}

for company, symbol in zip(companies, stock_symbols):
  question_debt = f"What is the total debt of {company} ({symbol}) in 2023 in US dollars?"
  documents = select_documents(stock_symbols=[symbol], form_types=["10-K"], fiscal_years=[2023])
  pages = retrieve_relevant_pages(question_debt, documents)
  total_debt = extract_value(question_debt, pages)
total_debts[company] = total_debt

# Step 2: Determine which company has the lowest total debt
lowest_debt_company = min(total_debts, key=total_debts.get)

# Step 3: Find the total revenues of that company
question_revenue = f"What is the total revenues of {lowest_debt_company} in 2023 in US dollars?"
symbol = stock_symbols[companies.index(lowest_debt_company)]
documents = select_documents(stock_symbols=[symbol], form_types=["10-K"], fiscal_years=[2023])
pages = retrieve_relevant_pages(question_revenue, documents)
total_revenues = extract_value(question_revenue, pages)

# Save the answer to a file
with open("answer.txt", "w" ) as f:
    f.write(total_revenues)

\end{lstlisting}
\caption{An example of code generated by GPT-4 in the CodeGen+DocS+PageR system. 
The LLM models successfully decompose the question into two main steps: 
(1) determine the company with the lowest total debt and 
(2) extract the Total Revenues of that company.
In the first step, the question is further decomposed into atomic questions for the Document Selection task and the Page Retrieval task.
This fine-grain process ensures the higher accuracy of the system compared to the coarse-grain systems such as Vanilla RAG.
}
\label{fig:codegen_code_example}
\end{figure*}

\newpage

\section{Example of question with metric description}

\begin{quote}
    \it 
    What is the Boeing's Total Debt in 2023?
    Where Total Debt is defined as:
    
    Total Debt is a supplemental line item across all templates with the following components: \\
    
    Standard template:
    
    Short-term Borrowings
    
    Current Portion of Long-Term Debt
    
    Current Portion of Leases
    
    Long-Term Debt
    
    Long Term Leases
    
    Finance Div. Debt Current
    
    Finance Div. Debt Non-Curr.\\
    
    Banks template:
    
    Short-Term Borrowings - (Bank Template)
    
    Current Portion of Long-Term Debt - (Bank Template)
    
    Current Portion of Leases
    
    Long-Term Debt
    
    Federal Home Loan Bank Debt - LT
    
    Long Term Leases
    
    Trust Pref. Securities\\
    
    All other templates:
    
    Short-Term Borrowings - (Template Specific)
    
    Curr. Port. of LT Debt
    
    Current Portion of Leases
    
    Long-Term Debt
    
    Long Term Leases
    
    Trust Pref. Securities
\end{quote}

\newpage

\begin{figure*}
\begin{lstlisting}[language=Python]
You are a financial expert.
The most current fiscal year is {current_year}
You can answer quantitative finance questions by writing Python code using helpful functions.
There are two functions:
- select_document: return a list of supporting documents.
- retrieve_relevant_pages: return a list of relevant pages that contain information to answer the question from the list of documents
- extract value: return an extracted value from the given document
select_document(
    companies: list = None,
    stock_symbols: list = None,
    form_types: list = None,
    fiscal_years: list = None,
    financial_period_end_date_range_start: str = None,
    financial_period_end_date_range_end: str = None
):
  """
  This function matches documents by a series of conditions.
  If the condition is not empty, they must match all given condition
  companies and stock_symbols are not mutually exclusive. A document is matched if satisfies one of these conditions.
  The documents must belong to one of the companies specified by the companies or stock_symbols
  The financial period end date to filter must be between (financial_period_end_date_range_start, financial_period_end_date_range_end)
  :param companies: a list of a few desired company short names.
  :param stock_symbols: a list of the corresponding companies' stock ticker symbols
  :param form_types: a list of the form types such as "8-K" for the current report, "10-K" for the annual report, "10-Q" for the quarterly report
  :param fiscal_years: a list of the corresponding companies' fiscal years
  :param financial_period_end_date_range_start: the beginning of a range used to filter financial period end date in "yyyy-mm-dd" format
  :param financial_period_end_date_range_end: the beginning of a range used to filter financial period end date in "yyyy-mm-dd" format

  :return: a list of supported documents. Return an empty list [] if no document is matched.
  """

retrieve_relevant_pages(question: str, documents: list):
  """
  :param question: a financial question
  :param documents: a list of documents, each with multiple pages
  :return: (list of str)a short list of pages
  """

extract_value(
  question: str,
  pages: list
):
  """
  :param qa: a question
  :param pages: a list of pages
  :return: an extracted value from the given list of pages.
  If the value is a money amount, the returned value is a float number in US dollars.
  If the value is a count, the returned value is a simple float number.
  If the value is a percentage, the returned value is a float number. E.g., 1% would be returned as 0.01.
  If the question is a yes-no question, it would return "Yes" or "No" only.
  """
Finally, you must write the short answer to a file named "answer.txt". The answer must be short, just a Yes/No, or a number

\end{lstlisting}
\caption{The prompt for CodeGen+DocS+PageR}
\label{fig:codegen_prompt}
\end{figure*}

\begin{figure*}
\begin{lstlisting}[language=Python]
Here are some examples:
Question: Did Coca-Cola pay dividends in 2017?
Python Code:
from functions import select_document
from functions import retrieve_relevant_pages
from functions import extract_value
documents = select_document(stock_symbols=["KO"], form_types=["10-K"], fiscal_years=[2017])
question = "How much did Coca-Cola pay dividends in USD in 2017?"
pages = retrieve_relevant_pages(question, documents)
value = extract_value(question, pages)
if isinstance(value, str):
    if value == "yes":
        answer = "Yes"
    elif value == "no":
        answer = "No"
    else:
        dividends = float(value)
else:
    dividends = float(value)
if dividends > 0:
    answer = "Yes"
else:
    answer = "No"
with open("answer.txt", "w") as f:
    f.write(answer)

Question: What is the overall revenue growth of Abbott over the last 2-year period?
Python Code:
from functions import select_document
from functions import retrieve_relevant_pages
from functions import extract_value
current_year = {current_year}
question = f"How much did Coca-Cola pay dividends in {{current_year}} in USD?"
documents = select_document(stock_symbols=["KO"], form_types=["10-K"], fiscal_years=[current_year])
pages = retrieve_relevant_pages(question, documents)
value_current = extract_value(question, pages=pages)
base_year = current_year - 2
question = f"How much did Coca-Cola pay dividends in {{base_year}} in USD?"
documents = select_document(stock_symbols=["KO"], form_types=["10-K"], fiscal_years=[base_year])
pages = retrieve_relevant_pages(question, documents)
value_base = extract_value(question, pages)
growth_percentage = (value_current - value_base) / value_base * 100.0
with open("answer.txt", "w") as f:
    f.write(str(growth_percentage))

Question: How much did NFLX return to the investors in the last 3 years?
Python Code:
from functions import select_document
from functions import retrieve_relevant_pages
from functions import extract_value
current_year = {current_year}
returned_values = []
for year in range(current_year, current_year - 3, -1):
question = f"How much did Netflix return to the investors in {{year}} in USD?"
    documents = select_document(companies=["Netflix"], stock_symbols=["NFLX"], form_types=["10-K"], fiscal_years=[year])
    pages = retrieve_relevant_pages(question, documents)
    return_in_us_dollars = extract_value(question, pages)
    returned_values.append(float(return_in_us_dollars))
total_return = sum(returned_values)
with open("answer.txt", "w") as f:
    f.write(str(total_return))
\end{lstlisting}
\caption{The few-shot demonstrations used for CodeGen+DocS+PageR system.}
\label{fig:codegen_few_shot_demonstrations}
\end{figure*}
\end{document}